\documentclass[preprint]{elsarticle}

\usepackage{hyperref}
\usepackage{algorithm}
\usepackage{algorithmic}
\usepackage{amsmath}
\usepackage{mathtools}
\usepackage{array}
\newcolumntype{P}[1]{>{\raggedright\arraybackslash}p{#1}}
\journal{Information Sciences}

\begin{document}

\begin{frontmatter}

\title{Distributed Correlation-Based Feature Selection in Spark}

\author[unah]{Raul-Jose Palma-Mendoza\corref{corrauthor}}
\ead{raul.palma@unah.edu.hn}

\author[uah]{Luis de-Marcos}
\ead{luis.demarcos@uah.es}

\author[uah]{Daniel Rodriguez}
\ead{daniel.rodriguezg@uah.es}

\author[udc]{Amparo Alonso-Betanzos}
\ead{amparo.alonso.betanzos@udc.es}

\cortext[corrauthor]{Corresponding author}

\address[unah]{Systems Engineering Department, \\
            National Autonomous University of Honduras. Blvd. Suyapa, Tegucigalpa, Honduras}
\address[uah]{Department of Computer Science, University of Alcal\'a \\
            Alcal\'a de Henares, 28871 Madrid, Spain}
\address[udc]{Department of Computer Science, University of A Coru\~na \\
            Campus de Elvi\~na s/n 15071 - A Coru\~na, Spain}

% %% Group authors per affiliation:
% \author{Elsevier\fnref{myfootnote}}
% \address{Radarweg 29, Amsterdam}
% \fntext[myfootnote]{Since 1880.}

% %% or include affiliations in footnotes:
% \author[mymainaddress,mysecondaryaddress]{Elsevier Inc}
% \ead[url]{www.elsevier.com}

% \author[mysecondaryaddress]{Global Customer Service\corref{mycorrespondingauthor}}
% \cortext[mycorrespondingauthor]{Corresponding author}
% \ead{support@elsevier.com}

% \address[mymainaddress]{1600 John F Kennedy Boulevard, Philadelphia}
% \address[mysecondaryaddress]{360 Park Avenue South, New York}

\begin{abstract}
Feature selection (FS) is a key preprocessing step in data mining. 
% However, traditional FS algorithms designed for execution on a single machine lack the necessary scalability to deal with the growing volumes of data becoming available in the current big data era. 
CFS (Correlation-Based Feature Selection) is an FS algorithm that has been successfully applied to classification problems in many domains. We describe Distributed CFS (DiCFS) as a completely redesigned, scalable, parallel and distributed version of the CFS algorithm, capable of dealing with the large volumes of data typical of big data applications. Two versions of the algorithm were implemented and compared using the Apache Spark cluster computing model, currently gaining popularity due to its much faster processing times than Hadoop's MapReduce model. We tested our algorithms on four publicly available datasets, each consisting of a large number of instances and two also consisting of a large number of features. 
%Our comprehensive experiments compared our versions with the classical non-distributed implementation of the CFS algorithm available in WEKA and with another recent Spark implementation of the CFS algorithm for regression problems.
The results show that our algorithms were superior in terms of both time-efficiency and scalability. In leveraging a computer cluster, they were able to handle larger datasets than the non-distributed WEKA version while maintaining the quality of the results, i.e., exactly the same features were returned by our algorithms when compared to the original algorithm available in WEKA.  
\end{abstract}


\end{frontmatter}

%\linenumbers

\section{Introduction}
\label{sec:intro}

In recent years, the advent of big data has raised unprecedented challenges for all types of organizations and researchers in many fields. Xindong et al. ~\cite{XindongWu2014}, however, state that the big data revolution has come to us not only with many challenges but also with plenty of opportunities for those organizations and researchers willing to embrace them. Data mining is one field where the opportunities offered by big data can be embraced, and, as indicated by Leskovec et al. ~\cite{Leskovec2014mining}, the main challenge is to extract useful information or knowledge from these huge data volumes that enable us to predict or better understand the phenomena involved in the generation of the data. 

Feature selection (FS) is a dimensionality reduction technique that has emerged as an important step in data mining. According to Guyon and Eliseeff~\cite{Guyon2003} its purpose is twofold: to select relevant attributes and simultaneously to discard redundant attributes. This purpose has become even more important nowadays, as vast quantities of data need to be processed in all kinds of disciplines. Practitioners also face the challenge of not having enough computational resources. In a review of the most widely used FS methods, Bol\'on-Canedo et al.~\cite{Bolon-Canedo2015b} conclude that there is a growing need for scalable and efficient FS methods, given that the existing methods are likely to prove inadequate for handling the increasing number of features encountered in big data.

Depending on their relationship with the classification process, FS methods are commonly classified in one of three main categories : (i) filter methods, (ii) wrapper methods, or (iii) embedded methods. \emph{Filters} rely solely on the characteristics of the data and, since they are independent of any learning scheme, they require less computational effort. They have been shown to be important preprocessing techniques, with many applications such as churn prediction~\cite{Idris2012,Idris2013} and microarray data classification. In microarray data classification, filters obtain better or at least comparable results in terms of accuracy to wrappers~\cite{Bolon-Canedo2015a}. In \emph{wrapper} methods, the final subset selection is based on a learning algorithm that is repeatedly trained with the data. Although wrappers tend to increase the final accuracy of the learning scheme, they are usually more computationally expensive than the other two approaches.  Finally, in \emph{embedded} methods, FS is part of the classification process, e.g., as happens with decision trees. 

Another important classification of FS methods is, according to their results, as (i) ranker algorithms or (ii) subset selector algorithms. With \emph{rankers}, the result is a sorted set of the original features. The order of this returned set is defined according to the quality that the FS method determines for each feature. Some rankers also assign a weight to each feature that provides more information about its quality. \emph{Subset selectors} return a non-ordered subset of features from the original set so that together they yield the highest possible quality according to some given measure. Subset selectors, therefore, consist of a search procedure and an evaluation measure. This can be considered an advantage in many cases, as rankers usually evaluate features individually and leave it to the user to select the number of top features in a ranking.

One filter-based subset selector method is the Correlation-Based Feature Selection (CFS) algorithm~\cite{Hall2000}, traditionally considered useful due to its ability not only to reduce dimensionality but also to improve classification algorithm performance. However, the CFS algorithm, like many other multivariate FS algorithms, has a time execution complexity $\mathcal{O}(m^2 \cdot n)$, where $m$ is the number of features and $n$ is the number of instances. This quadratic complexity in the number of features makes CFS very sensitive to the \textit{the curse of dimensionality}~\cite{bellman1957dynamic}. Therefore, a scalable adaptation of the original algorithm is required to be able to apply the CFS algorithm to datasets that are large both in number of instances and dimensions.

As a response to the big data phenomenon, many technologies and programming frameworks have appeared with the aim of helping data mining practitioners design new strategies and algorithms that can tackle the challenge of distributing work over clusters of computers. One such tool that has recently received much attention is Apache Spark~\cite{Zaharia2010}, which represents a new programming model that is a superset of the MapReduce model introduced by Google~\cite{Dean2004a,Dean2008}. One of Spark's strongest advantages over the traditional MapReduce model is its ability to efficiently handle the iterative algorithms that frequently appear in the data mining and machine learning fields.

We describe two distributed and parallel versions of the original CFS algorithm for classification problems using the Apache Spark programming model. The main difference between them is how the data is distributed across the cluster, i.e., using a horizontal partitioning scheme (hp) or using a vertical partitioning scheme (vp). We compare the two versions -- DiCFS-hp  and DiCFS-vp, respectively -- and also compare them with a baseline, represented by the classical non-distributed implementation of CFS in WEKA~\cite{Hall2009a}. Finally, their benefits in terms of reduced execution time are compared with those of the CFS version developed by Eiras-Fanco et al.~\cite{Eiras-Franco2016} for regression problems. The results show that the time-efficiency and scalability of our two versions are an improvement on those of the original version of the CFS; furthermore, similar or improved execution times are obtained with respect to the Eiras-Franco et al~\cite{Eiras-Franco2016} regression version. In the interest of reproducibility, our software and sources are available as a Spark package\footnote{\url{https://spark-packages.org}} called DiCFS, with a corresponding mirror in Github.\footnote{\url{https://github.com/rauljosepalma/DiCFS}}

The rest of this paper is organized as follows. Section~\ref{sec:stateofart} summarizes the most important contributions in the area of distributed and parallel FS and proposes a classification according to how parallelization is carried out. Section~\ref{sec:cFS} describes the original CFS algorithm, including its theoretical foundations. Section~\ref{sec:spark} presents the main aspects of the Apache Spark computing framework, focusing on those relevant to the design and implementation of our proposed algorithms. Section~\ref{sec:diCFS} describes and discusses our DiCFS-hp and DiCFS-vp versions of the CFS algorithm.  Section~\ref{sec:experiments} describes our experiments to compare results for DiCFS-hp and DiCFS-vp, the WEKA approach and the Eiras-Fanco et al.~\cite{Eiras-Franco2016} approach. Finally, conclusions and future work are outlined in Section~\ref{sec:conclusions}.

\section{Background and Related Work}
\label{sec:stateofart}

As might be expected, filter-based FS algorithms have asymptotic complexities that depend on the number of features and/or instances in a dataset. Many algorithms, such as the CFS, have quadratic complexities, while the most frequently used algorithms have at least linear complexities~\cite{Bolon-Canedo2015b}. This is why, in recent years, many attempts have been made to achieve more scalable FS methods. In what follows, we analyse recent work on the design of new scalable FS methods according to parallelization approaches: (i) search-oriented, (ii) dataset-split-oriented, or (iii) filter-oriented.

\emph{Search-oriented} parallelizations account for most approaches, in that the main aspects to be parallelized are (i) the search guided by a classifier and (ii) the corresponding evaluation of the resulting models. We classify the following studies in this category:

\begin{itemize}

\item Kubica et al.~\cite{Kubica2011} developed parallel versions of three forward-search-based FS algorithms, where a wrapper with a logistic regression classifier is used to guide a search parallelized using the MapReduce model. 

\item Garc\'ia et al.~\cite{Garcia_aparallel} presented a simple approach for parallel FS, based on selecting random feature subsets and evaluating them in parallel using a classifier. In their experiments they used a support vector machine (SVM) classifier and, in comparing their results with those for a traditional wrapper approach, found lower accuracies but also much shorter computation times. 

\item Wang et al.~\cite{Wang2016} used the Spark computing model to implement an FS strategy for classifying network traffic. They first implemented an initial FS using the Fisher score filter~\cite{duda2012pattern} and then performed, using a wrapper approach, a distributed forward search over the best $m$ features selected. Since the Fisher filter was used, however, only numerical features could be handled.

\item Silva et al.~\cite{Silva2017} addressed the FS scaling problem using an asynchronous search approach, given that synchronous search, as commonly performed, can lead to efficiency losses due to the inactivity of some processors waiting for other processors to end their tasks. In their tests, they first obtained an initial reduction using a mutual information (MI)~\cite{Peng2005} filter and then evaluated subsets using a random forest (RF)~\cite{Ho1995} classifier. However, as stated by those authors, any other approach could be used for subset evaluation.

\end{itemize}

\emph{Dataset-split-oriented} approaches have the main characteristic that parallelization is performed by splitting the dataset vertically or horizontally, then applying existing algorithms to the parts and finally merging the results following certain criteria. We classify the following studies in this category:

\begin{itemize}

\item Peralta et al.~\cite{Peralta2015} used the MapReduce model to implement a wrapper-based evolutionary search FS method. The dataset was split by instances and the FS method was applied to each resulting subset. Simple majority voting was used as a reduction step for the selected features and the final subset of feature was selected according to a user-defined threshold. All tests were carried out using the EPSILON dataset, which we also use here (see Section~\ref{sec:experiments}).

\item Bol\'on-Canedo et al.~\cite{Bolon-Canedo2015a} proposed a framework to deal with high dimensionality data by first optionally ranking features using a FS filter, then partitioning vertically by dividing the data according to features (columns) rather than, as commonly done, according to instances (rows). After partitioning, another FS filter is applied to each partition, and finally, a merging procedure guided by a classifier obtains a single set of features. The authors experiment with five commonly used FS filters for the partitions, namely, CFS~\cite{Hall2000}, Consistency~\cite{Dash2003}, INTERACT~\cite{Zhao2007}, Information Gain~\cite{Quinlan1986} and ReliefF~\cite{Kononenko1994}, and with four classifiers for the final merging, namely, C4.5~\cite{Quinlan1992}, Naive Bayes~\cite{rish2001empirical}, $k$-Nearest Neighbors~\cite{Aha1991} and SVM~\cite{vapnik1995nature}, show that their own approach significantly reduces execution times while maintaining and, in some cases, even improving accuracy.

\end{itemize}

Finally, \emph{filter-oriented} methods include redesigned or new filter methods that are, or become, inherently parallel. Unlike the methods in the other categories, parallelization in this category methods can be viewed as an internal, rather than external, element of the algorithm. We classify the following studies in this category:

\begin{itemize}
\item Zhao et al.~\cite{Zhao2013a} described a distributed parallel FS method based on a variance preservation criterion using the proprietary software SAS High-Performance Analytics.~\footnote{\url{http://www.sas.com/en_us/software/high-performance-analytics.html}} One remarkable characteristic of the method is its support not only for supervised FS, but also for unsupervised FS where no label information is available. Their experiments were carried out with datasets with both high dimensionality and a high number of instances. 

\item Ram\'irez-Gallego et al.~\cite{Ramirez-Gallego2017} described scalable versions of the popular mRMR~\cite{Peng2005} FS filter that included a distributed version using Spark. The authors showed that  their version that leveraged the power of a cluster of computers could perform much faster than the original and processed much larger datasets.

\item In a previous work~\cite{Palma-Mendoza2018}, using the Spark computing model we designed a distributed version of the ReliefF~\cite{Kononenko1994} filter, called DiReliefF. In testing using datasets with large numbers of features and instances, it was much more efficient and scalable than the original filter.

\item Finally, Eiras-Franco et al~\cite{Eiras-Franco2016}, using four distributed FS algorithms, three of them filters, namely, InfoGain~\cite{Quinlan1986}, ReliefF~\cite{Kononenko1994} and the CFS~\cite{Hall2000}, reduce execution times with respect to the original versions. However, in the CFS case, the version of those authors focuses on regression problems where all the features, including the class label, are numerical, with correlations calculated using the Pearson coefficient. A completely different approach is required to design a parallel version for classification problems where correlations are based on the information theory.
\end{itemize}

The approach described here can be categorized as a \emph{filter-oriented} approach that builds on works described elsewhere~\cite{Ramirez-Gallego2017},~\cite{Palma-Mendoza2018}, ~\cite{Eiras-Franco2016}. The fact that their focus was not only on designing an efficient and scalable FS algorithm, but also on preserving the original behaviour (and obtaining the same final results) of traditional filters, means that research focused on those filters is also valid for adapted versions. Another important issue in relation to filters is that, since they are generally more efficient than wrappers, they are often the only feasible option due to the abundance of data. It is worth mentioning that scalable filters could feasibly be included in any of the methods mentioned in the \emph{search-oriented} and \emph{dataset-split-oriented} categories, where an initial filtering step is implemented to improve performance. 

\section{Correlation-Based Feature Selection (CFS)}
\label{sec:cFS}
 
The CFS method, originally developed by Hall~\cite{Hall2000}, is categorized as a subset selector because it evaluates subsets rather than individual features. For this reason, the CFS needs to perform a search over candidate subsets, but since performing a full search over all possible subsets is prohibitive (due to the exponential complexity of the problem), a heuristic has to be used to guide a partial search. This heuristic is the main concept behind the CFS algorithm, and, as a filter method, the CFS is not a classification-derived measure, but rather applies a principle derived from Ghiselly's test theory~\cite{ghiselli1964theory}, i.e., \emph{good feature subsets contain features highly correlated with the class, yet uncorrelated with each other}.

This principle is formalized in Equation~(\ref{eq:heuristic}) where $M_s$ represents the merit assigned by the heuristic to a subset $s$ that contains $k$ features, $\overline{r_{cf}}$ represents the average of the correlations between each feature in $s$ and the class attribute, and $\overline{r_{ff}}$ is the average correlation between each of the $\begin{psmallmatrix}k\\2\end{psmallmatrix}$ possible feature pairs in $s$. The numerator can be interpreted as an indicator of how predictive the feature set is and the denominator can be interpreted as an indicator of how redundant features in $s$ are.

\begin{equation}
\label{eq:heuristic}
M_s = \frac { k\cdot \overline { r_{cf} }  }{ \sqrt { k + k (k - 1) \cdot \overline{ r_{ff}} } } 
\end{equation}

Equation~(\ref{eq:heuristic}) also posits the second important concept underlying the CFS, which is the computation of correlations to obtain the required averages. In classification problems, the CFS uses the symmetrical uncertainty (SU) measure~\cite{press1982numerical} shown in Equation~(\ref{eq:su}), where $H$ represents the entropy function of a single or conditioned random variable, as shown in Equation~(\ref{eq:entropy}). This calculation adds a requirement for the dataset before processing, which is that all non-discrete features must be discretized. By default, this process is performed using the discretization algorithm proposed by Fayyad and Irani~\cite{Fayyad1993}.

\begin{equation}
\label{eq:su}
SU = 2 \cdot \left[ \frac { H(X) - H(X|Y) }{ H(Y) + H(X) }  \right]  
\end{equation}

\begin{align}
\label{eq:entropy}
H(X) &=-\sum _{ x\in X }{ p(x)\log _{2}{p(x)} } \nonumber \\
H(X | Y) &=-\sum _{ y\in Y }{ p(y) } \sum_{x \in X}{p(x |y) \log _{ 2 }{ p(x | y) } } 
\end{align}

The third core CFS concept is its search strategy. By default, the CFS algorithm uses a best-first search to explore the search space. The algorithm starts with an empty set of features and at each step of the search all possible single feature expansions are generated. The new subsets are evaluated using Equation~(\ref{eq:heuristic}) and are then added to a priority queue according to merit. In the subsequent iteration, the best subset from the queue is selected for expansion in the same way as was done for the first empty subset. If expanding the best subset fails to produce an improvement in the overall merit, this counts as a \emph{fail} and the next best subset from the queue is selected. By default, the CFS uses five consecutive fails as a stopping criterion and as a limit on queue length.

The final CFS element is an optional post-processing step. As stated before, the CFS tends to select feature subsets with low redundancy and high correlation with the class. However, in some cases, extra features that are \emph{locally predictive} in a small area of the instance space may exist that can be leveraged by certain classifiers~\cite{Hall1999}. To include these features in the subset after the search, the CFS can optionally use a heuristic that enables inclusion of all features whose correlation with the class is higher than the correlation between the features themselves and with features already selected. Algorithm~\ref{alg:cFS} summarizes the main aspects of the CFS.

\begin{algorithm}
\caption{CFS~\cite{Hall2000}}
\label{alg:cFS}
\begin{algorithmic}[1] 
\STATE $Corrs := $ correlations between all features with the class \label{lin:allCorrs}
\STATE $BestSubset := \emptyset$
\STATE $Queue.setCapacity(5)$
\STATE $Queue.add(BestSubset)$
\STATE $NFails := 0$
\WHILE{$NFails < 5$}
  \STATE $HeadState := Queue.dequeue$ \COMMENT {Remove from queue}
  \STATE $NewSubsets := evaluate(expand(HeadState), Corrs)$ \label{lin:expand}
  \STATE $Queue.add(NewSubsets)$
  \IF{$Queue.isEmpty$}
    \RETURN $BestSubset$ \COMMENT {When the best subset is the full subset}
  \ENDIF
  \STATE $LocalBest := Queue.head$ \COMMENT {Check new best without removing}
  \IF{$LocalBest.merit > BestSubset.merit$}
    \STATE $BestSubset := LocalBest$ \COMMENT{Found a new best}
    \STATE $NFails := 0$ \COMMENT{Fails must happen consecutively}
  \ELSE
    \STATE $NFails := NFails + 1$
  \ENDIF
\ENDWHILE
\STATE \COMMENT {Optionally add locally predictive features to $BestSubset$}
\RETURN $BestSubset$
\end{algorithmic}
\end{algorithm}

\section{The Spark Cluster Computing Model}
\label{sec:spark}

The following short description of the main concepts behind the Spark computing model focuses exclusively on aspects that complete the conceptual basis for our DiCFS proposal in Section~\ref{sec:diCFS}. 

The main concept behind the Spark model is what is known as the resilient distributed dataset (RDD). Zaharia et al.~\cite{Zaharia2010,Zaharia2012} defined an RDD as a read-only collection of objects, i.e., a dataset partitioned and distributed across the nodes of a cluster. The RDD has the ability to automatically recover lost partitions through a lineage record that knows the origin of the data and possible calculations done. Even more relevant for our purposes is the fact that operations run for an RDD are automatically parallelized by the Spark engine; this abstraction frees the programmer from having to deal with threads, locks and all other complexities of traditional parallel programming.

With respect to the cluster architecture, Spark follows the master-slave model. Through a cluster manager (master), a driver program can access the cluster and coordinate the execution of a user application by assigning tasks to the executors, i.e., programs that run in worker nodes (slaves). By default, only one executor is run per worker. Regarding the data, RDD partitions are distributed across the worker nodes, and the number of tasks launched by the driver for each executor is set according to the number of RDD partitions residing in the worker. 

Two types of operations can be executed on an RDD, namely, actions and transformations. Of the \emph{actions}, which allow results to be obtained from a Spark cluster, perhaps the most important is $collect$, which returns an array with all the elements in the RDD. This operation has to be done with care, to avoid exceeding the maximum memory available to the driver. Other important actions include $reduce$, $sum$, $aggregate$ and $sample$, but as they are not used by us here, we will not explain them. \emph{Transformations} are mechanisms for creating an RDD from another RDD. Since RDDs are read-only, a transformation creating a new RDD does not affect the original RDD. A basic transformation is $mapPartitions$, which receives, as a parameter, a function that can handle all the elements of a partition and return another collection of elements to conform a new partition. The $mapPartitions$ transformation is applied to all partitions in the RDD to obtain a new transformed RDD. Since received and returned  partitions do not need to match in size, $mapPartitions$ can thus reduce or increase the overall size of an RDD. Another interesting transformation is $reduceByKey$; this can only be applied to what is known as a $PairRDD$, which is an RDD whose elements are key-value pairs, where the keys do not have to be unique. The $reduceByKey$ transformation is used to aggregate the elements of an RDD, which it does by applying a commutative and associative function that receives two values of the PairRDD as arguments and returns one element of the same type. This reduction is applied by key, i.e., elements with the same key are reduced such that the final result is a PairRDD with unique keys, whose corresponding values are the result of the reduction. Other important transformations (which we do not explain here) are $map$, $flatMap$ and $filter$.

Another key concept in Spark is \emph{shuffling}, which refers to the data communication required for certain types of transformations, such as the above-mentioned $reduceByKey$. Shuffling is a costly operation because it requires redistribution of the data in the partitions, and therefore, data read and write across all nodes in the cluster. For this reason, shuffling operations are minimized as much as possible.

The final concept underpinning our proposal is \emph{broadcasting}, which is a useful mechanism for efficiently sharing read-only data between all worker nodes in a cluster. Broadcast data is dispatched from the driver throughout the network and is thus made available to all workers in a deserialized fast-to-access form.

\section{Distributed Correlation-Based Feature Selection (DiCFS)}
\label{sec:diCFS}

We now describe the two algorithms that conform our proposal. They represent alternative distributed versions that use different partitioning strategies to process the data. We start with some considerations common to both approaches.

As stated previously, CFS has a time execution complexity of $\mathcal{O}(m^2 \cdot n)$ where $m$ is the number of features and $n$ is the number of instances. This complexity derives from the first step shown in Algorithm~\ref{alg:cFS}, the calculation of $\begin{psmallmatrix}m+ 1\\2\end{psmallmatrix}$ correlations between all pairs of features including the class, and the fact that for each pair, $\mathcal{O}(n)$ operations are needed in order to calculate the entropies. Thus, to develop a scalable version, our main focus in parallelization design must be on the calculation of correlations.

Another important issue is that, although the original study by Hall~\cite{Hall2000} stated that all correlations had to be calculated before the search, this is only a true requisite when a backward best-first search is performed. In the case of the search shown in Algorithm~\ref{alg:cFS}, correlations can be calculated on demand, i.e., on each occasion a new non-evaluated pair of features appears during the search. In fact, trying to calculate all correlations in any dataset with a high number of features and instances is prohibitive; the tests performed on the datasets described in Section~\ref{sec:experiments} show that a very low percentage of correlations is actually used during the search and also that on-demand correlation calculation is around $100$ times faster when the default number of five maximum fails is used.

Below we describe our two alternative methods for calculating these correlations in a distributed manner depending on the type of partitioning used.

\subsection{Horizontal Partitioning}
\label{subsec:horizontalPart}

Horizontal partitioning of the data may be the most natural way to distribute work between the nodes of a cluster. If we consider the default layout where the data is represented as a matrix $D$ in which the columns represent the different features and the rows represent the instances, then it is natural to distribute the matrix by assigning different groups of rows to nodes in the cluster. If we represent this matrix as an RDD, this is exactly what Spark will automatically do.

Once the data is partitioned, Algorithm~\ref{alg:cFS} (omitting line~\ref{lin:allCorrs}) can be started on the driver. The distributed work will be performed on line~\ref{lin:expand}, where the best subset in the queue is expanded and, depending on this subset and the state of the search, a number $nc$ of new pairs of correlations will be required to evaluate the resulting subsets. Thus, the most complex step is the calculation of the corresponding $nc$ contingency tables that will allow us to obtain the entropies and conditional entropies that conform the symmetrical uncertainty correlation (see Equation~(\ref{eq:su})). These $nc$ contingency tables are partially calculated locally by the workers following Algorithm~\ref{alg:localCTables}. As can be observed, the algorithm loops through all the local rows, counting the values of the features contained in \emph{pairs} (declared in line~\ref{lin:pairs}) and storing the results in a map holding the feature pairs as keys and the contingency tables as their matching values.

The next step is to merge the contingency tables from all the workers to obtain global results. Since these tables hold simple value counts, they can easily be aggregated by performing an element-wise sum of the corresponding tables. These steps are summarized in Equation~(\ref{eq:cTables}), where $CTables$ is an RDD of keys and values, and where each key corresponds to a feature pair and each value to a contingency table.

\begin{algorithm}
\caption{function localCTables(pairs)(partition)}
\label{alg:localCTables}
\begin{algorithmic}[1]
\STATE $pairs \leftarrow$ $nc$ pairs of features \label{lin:pairs}
\STATE $rows \leftarrow$ local rows of $partition$
\STATE $m \leftarrow$ number of columns (features in $D$)
\STATE $ctables \leftarrow$ a map from each pair to an empty contingency table
\FORALL{$r \in rows$}
  \FORALL{ $(x,y) \in pairs$ }
    \STATE $ctables(x,y)(r(x),r(y))$ += $1$
  \ENDFOR
\ENDFOR
\RETURN $ctables$
\end{algorithmic}
\end{algorithm}

\begin{align}
\label{eq:cTables}
pairs &= \left \{ (feat_a, feat_b), \cdots, (feat_x, feat_y)  \right \} \nonumber \\
nc &= \left | pairs \right | \nonumber \\
CTables &= D.mapPartitions(localCTables(pairs)).reduceByKey(sum) \nonumber \\
CTables &= 
\begin{bmatrix}
((feat_a, feat_b), ctable_{a,b})\\ 
\vdots \\ 
((feat_x, feat_y), ctable_{x,y})\\ 
\end{bmatrix}_{nc \times 1} \nonumber \\
\end{align}

Once the contingency tables have been obtained, the calculation of the entropies and conditional entropies is straightforward since all the information necessary for each calculation is contained in a single row of the $CTables$ RDD. This calculation can therefore be performed in parallel by processing the local rows of this RDD.

Once the distributed calculation of the correlations is complete, control returns to the driver, which continues execution of line~\ref{lin:expand} in Algorithm~\ref{alg:cFS}. As can be observed, the distributed work only happens when new correlations are needed, and this occurs in only two cases: (i) when new pairs of features need to be evaluated during the search, and (ii) at the end of the execution if the user requests the addition of locally predictive features. 

\begin{figure}
  \includegraphics[width=1\textwidth]{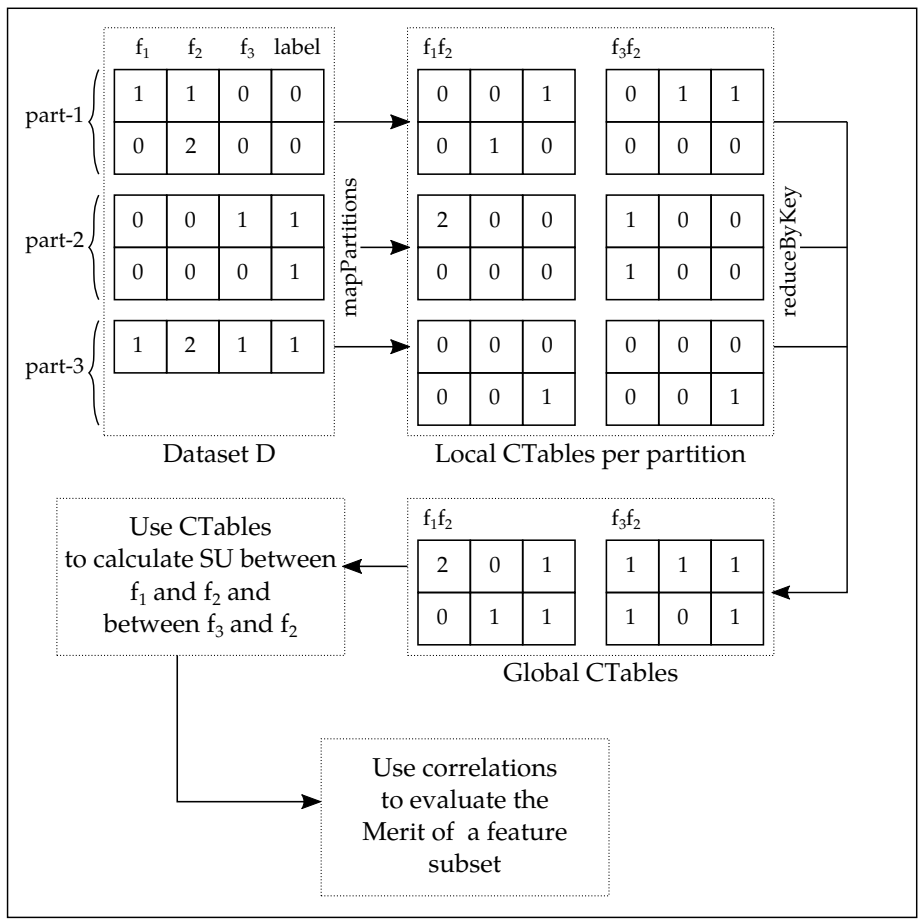}
\caption{Horizontal partitioning steps for a small dataset D to obtain the correlations needed to evaluate a features subset}
\label{fig:horizontalPartResume}
\end{figure}

To sum up, every iteration in Algorithm~\ref{alg:cFS} expands the current best subset and obtains a group of subsets for evaluation. This evaluation requires a merit, and the merit for each subset is obtained according to Figure~\ref{fig:horizontalPartResume}, which illustrates the most important steps in the horizontal partitioning scheme using a case where correlations between features f2 and f1 and between f2 and f3 are calculated in order to evaluate a subset.

\subsection{Vertical Partitioning}
\label{subsec:vecticalPart}

\begin{figure}
  \includegraphics[width=1\textwidth]{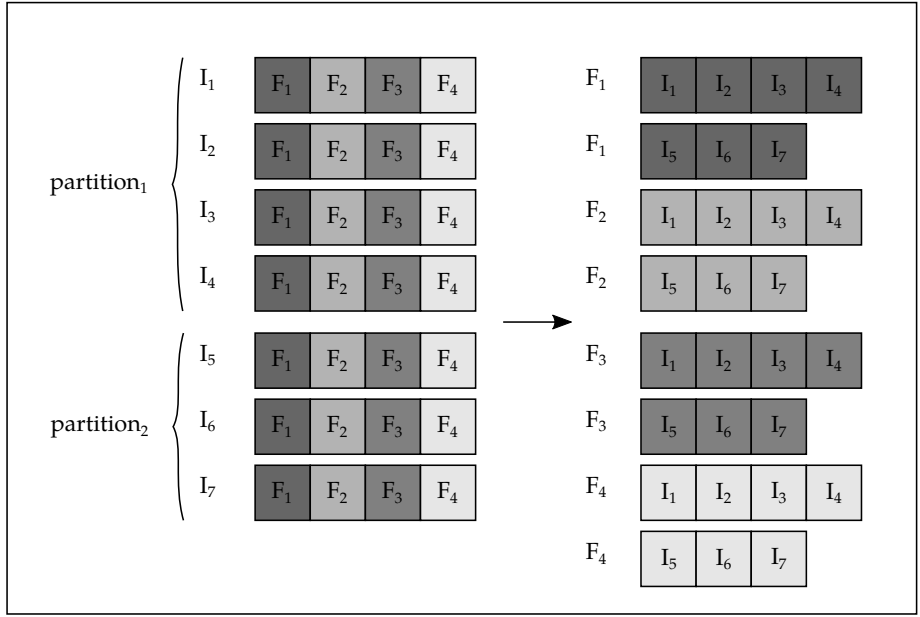}
\caption{Example of a columnar transformation of a small dataset with two partitions, seven instances and four features (from~\cite{Ramirez-Gallego2017})}
\label{fig:columnarTrans}
\end{figure}

Vertical partitioning has already been proposed in Spark by Ram\'irez-Gallego et al.~\cite{Ramirez-Gallego2017}, using another important FS filter, mRMR. Although mRMR is a ranking algorithm (it does not select subsets), it also requires the calculation of information theory measures such as entropies and conditional entropies between features. Since data is distributed horizontally by Spark, those authors propose two main operations to perform the vertical distribution:

\begin{itemize}
  \item \emph{Columnar transformation}. Rather than use the traditional format whereby the dataset is viewed as a matrix whose columns represent features and  rows represent instances, a transposed version is used in which the data represented as an RDD is distributed by features and not by instances, in such a way that the data for a specific feature will in most cases be stored and processed by the same node. Figure~\ref{fig:columnarTrans}, based on Ram\'irez-Gallego et al.~\cite{Ramirez-Gallego2017}, explains the process using an example based on a dataset with two partitions, seven instances and four features.
  
  \item \emph{Feature broadcasting}. Because features must be processed in pairs to calculate conditional entropies and because different features can be stored in different nodes, some features are broadcast over the cluster so all nodes can access and evaluate them along with the other stored features.
\end{itemize}

In the case of the adapted mRMR~\cite{Ramirez-Gallego2017}, since every step in the search requires the comparison of a single feature with a group of remaining features, it proves efficient, at each step, to broadcast this single feature (rather than multiple features). In the case of the CFS, the core issue is that, at any point in the search when expansion is performed, if the size of subset being expanded is $k$, then the correlations between the $m-k$ remaining features and $k-1$ features in the subset being expanded have already been calculated in previous steps; consequently, only the correlations between the most recently added feature and the $m-k$ remaining features are missing. Therefore, the proposed operations can  be applied efficiently in the CFS just by broadcasting the most recently added feature.

The disadvantages of vertical partitioning are that (i) it requires an extra processing step to change the original layout of the data and this requires shuffling, (ii) it needs data transmission to broadcast a single feature in each search step, and (iii) the fact that, by default, the dataset is divided into a number of partitions equal to the number of features $m$ in the dataset may not be optimal for all cases (while this parameter can be tuned, it can never exceed $m$). The main advantage of vertical partitioning is that the data layout and the broadcasting of the compared feature move all the information needed to calculate the contingency table to the same node, which means that this information can be more efficiently processed locally. Another advantage is that the whole dataset does not need to be read every time a new set of features has to be compared, since the dataset can be filtered by rows to process only the required features. 

Due to the nature of the search strategy (best-first) used in the CFS, the first search step will always involve all features, so no filtering can be performed. For each subsequent step, only one more feature per step can be filtered out. This is especially important with high dimensionality datasets: the fact that the number of features is much higher than the number of search steps means that the percentage of features that can be filtered out is reduced.

We performed a number of experiments to quantify the effects of the advantages and disadvantages of each approach and to check the conditions in which one approach was better than the other.

\section{Experiments}
\label{sec:experiments}

The experiments tested and compared time-efficiency and scalability for the horizontal and vertical DiCFS approaches so as to check whether they improved on the original non-distributed version of the CFS. We also tested and compared execution times with that reported in the recently published research by Eiras-Franco et al.~\cite{Eiras-Franco2016} into a distributed version of CFS for regression problems. 

Note that no experiments were needed to compare the quality of the results for the distributed and non-distributed CFS versions as the distributed versions were designed to return the same results as the original algorithm.

For our experiments, we used a single master node and up to ten slave nodes from the big data platform of the Galician Supercomputing Technological Centre (CESGA).~\footnote{\url{http://bigdata.cesga.es/}} The nodes have the following configuration:

\begin{itemize}
\item CPU: 2 X Intel Xeon E5-2620 v3 @ 2.40GHz 
\item CPU Cores: 12 (2X6)
\item Total Memory: 64 GB
\item Network: 10GbE
\item Master Node Disks: 8 X 480GB SSD SATA 2.5" MLC G3HS
\item Slave Node Disks: 12 X 2TB NL SATA 6Gbps 3.5" G2HS
\item Java version: OpenJDK 1.8
\item Spark version: 1.6
\item Hadoop (HDFS) version: 2.7.1
\item WEKA version: 3.8.1
\end{itemize}

The experiments were run on four large-scale publicly available datasets. The ECBDL14~\cite{Bacardit2012} dataset, from the protein structure prediction field, was used in the ECBLD14 Big Data Competition included in the GECCO'2014 international conference. This  dataset has approximately 33.6 million instances, 631 attributes and 2 classes, consists 98\% of negative examples and occupies about 56GB of disk space. HIGGS~\cite{Sadowski2014}, from the UCI Machine Learning Repository~\cite{Lichman2013}, is a recent dataset  representing a classification problem that distinguishes between a signal process which produces Higgs bosons and a background process which does not. KDDCUP99~\cite{Ma2009} represents data from network connections and classifies them as normal connections or different types of attacks (a multi-class problem). Finally, EPSILON is an artificial dataset built for the Pascal Large Scale Learning Challenge in 2008.\footnote{\url{http://largescale.ml.tu-berlin.de/about/}} Table~\ref{tbl:datasets} summarizes the main characteristics of the datasets.

\begin{table}[tbp]
% \centering
\caption{Description of the four datasets used in the experiments}
\label{tbl:datasets}
\begin{tabular}{P{1in}P{0.7in}P{0.7in}P{0.7in}P{0.7in}}
\hline
Dataset                      & No. of Samples ($\times 10^{6}$) & No. of Features. & Feature Types          & Problem Type \\ \hline
ECBDL14~\cite{Bacardit2012}  & $\sim$33.6                       & 632              & Numerical, Categorical & Binary       \\
HIGGS~\cite{Sadowski2014}    & 11                               & 28               & Numerical              & Binary       \\
KDDCUP99~\cite{Ma2009}       & $\sim$5                          & 42               & Numerical, Categorical & Multiclass   \\
EPSILON                      & 1/2                              & 2,000            & Numerical              & Binary       \\ \hline
\end{tabular}
\end{table}

With respect to algorithm parameter configuration, two defaults were used in all the experiments: the inclusion of locally predictive features and the use of five consecutive fails as a stopping criterion. These defaults apply to both distributed and non-distributed versions. Moreover, for the vertical partitioning version, the number of partitions was equal to the number of features, as set by default in Ram\'irez-Gallego et al.~\cite{Ramirez-Gallego2017}. The horizontally and vertically distributed versions of the CFS are labelled DiCFS-hp and DiCFS-vp, respectively.

\begin{figure}
  \includegraphics[width=1\textwidth]{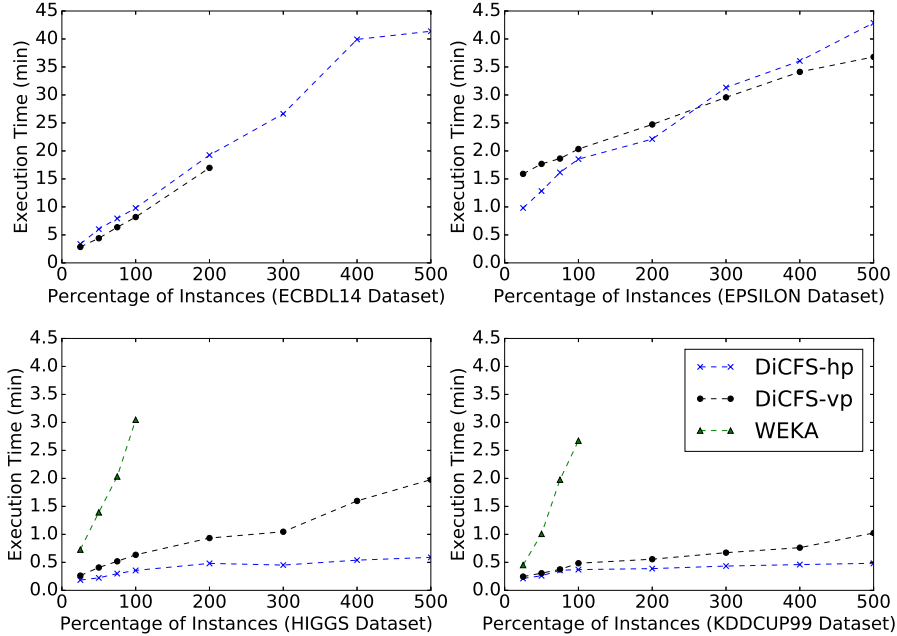}
\caption{Execution time with respect to percentages of instances in four datasets, for DiCFS-hp and DiCFS-vp using ten nodes and for a non-distributed implementation in WEKA using a single node}
\label{fig:execTimeVsNInsta}
\end{figure}

We first compared execution times for the four algorithms in the datasets using ten slave nodes with all their cores available. For the case of the non-distributed version of the CFS, we used the implementation provided in the WEKA platform~\cite{Hall2009a}. The results are shown in Figure~\ref{fig:execTimeVsNInsta}.

Note that, with the aim of offering a comprehensive view of execution time behaviour, Figure~\ref{fig:execTimeVsNInsta} shows results for sizes larger than the 100\% of the datasets. To achieve these sizes, the instances in each dataset were duplicated as many times as necessary. Note also that, since  ECBDL14 is a very large dataset, its temporal scale is different from that of the other datasets.

Regarding the non-distributed version of the CFS, Figure~\ref{fig:execTimeVsNInsta} does not show results for WEKA in the experiments on the ECBDL14 dataset, because it was impossible to execute that version in the CESGA platform due to memory requirements exceeding the available limits. This also occurred with the larger samples from the EPSILON dataset for both algorithms: DiCFS-vp and DiCFS-hp. Even when it was possible to execute the WEKA version with the two smallest samples from the EPSILON dataset, these samples are not shown because the execution times were too high (19 and 69 minutes, respectively). Figure~\ref{fig:execTimeVsNInsta} shows successful results for the smaller HIGGS and KDDCUP99 datasets, which could still be processed in a single node of the cluster, as required by the non-distributed version. However, even in the case of these smaller datasets, the execution times of the WEKA version were worse compared to those of the distributed versions.

Regarding the distributed versions, DiCFS-vp was unable to process the oversized versions of the ECBDL14 dataset, due to the large amounts of memory required to perform shuffling. The HIGGS and KDDCUP99 datasets showed an increasing difference in favor of DiCFS-hp, however, due to the fact that these datasets have much smaller feature sizes than ECBDL14 and EPSILON. As mentioned earlier, DiCFS-vp ties parallelization to the number of features in the dataset, so datasets with small numbers of features were not able to fully leverage the cluster nodes. Another view of the same issue is given by the results for the EPSILON dataset; in this case, DiCFS-vp obtained the best execution times for the 300\% sized and larger datasets. This was because there were too many partitions (2,000) for the number of instances available in smaller than 300\% sized datasets; further experiments showed that adjusting the number of partitions to 100 reduced the execution time of DiCFS-vp for the 100\% EPSILON dataset from about 2 minutes to 1.4 minutes (faster than DiCFS-hp). Reducing the number of partitions further, however, caused the execution time to start increasing again.

\begin{figure}
  \includegraphics[width=1\textwidth]{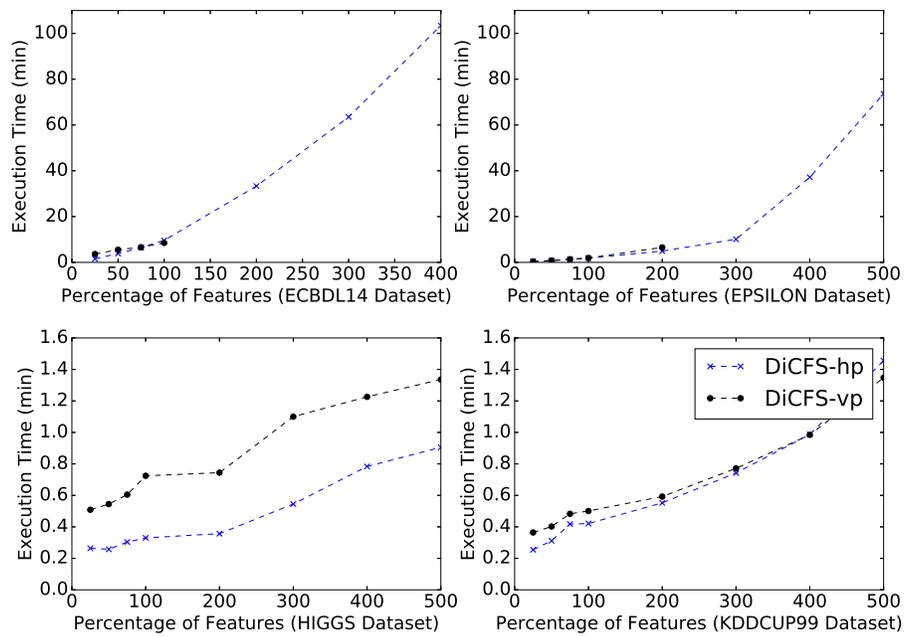}
\caption{Execution times with respect to different percentages of features in four datasets for DiCFS-hp and DiCFS-vp}
\label{fig:execTimeVsNFeats}
\end{figure}

Figure~\ref{fig:execTimeVsNFeats} shows the results for similar experiments, except that this time the percentage of features in the datasets was varied and the features were copied to obtain oversized versions of the datasets. It can be observed that the number of features had a greater impact on the memory requirements of DiCFS-vp. This caused problems not only in processing the ECBDL14 dataset but also the EPSILON dataset. We can also see quadratic time complexity in the number of features and how the temporal scale in the EPSILON dataset (with the highest number of dimensions) matches that of the ECBDL14 dataset. As for the KDDCUP99 dataset, the results show that increasing the number of features obtained a better level of parallelization and a slightly improved execution time of DiCFS-vp compared to DiCFS-hp for the 400\% dataset version and above.

\begin{figure}
  \includegraphics[width=1\textwidth]{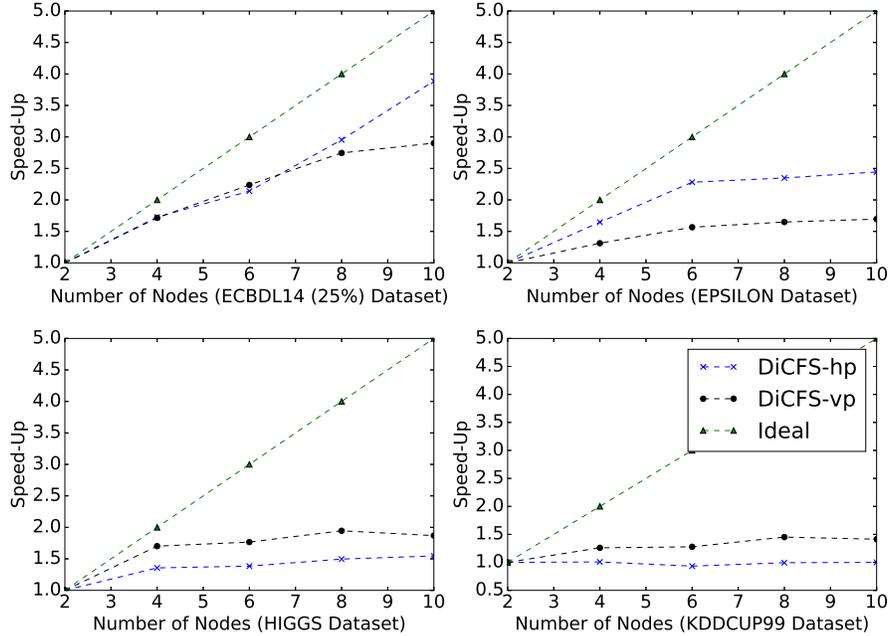}
\caption{Speed-up for four datasets for DiCFS-hp and DiCFS-vp}
\label{fig:speedup}
\end{figure}

An important measure of the scalability of an algorithm is \emph{speed-up}, which is a measure that indicates how capable an algorithm is of leveraging a growing number of nodes so as to reduce execution times. We used the speed-up definition shown in Equation~(\ref{eq:speedup}) and used all the available cores for each node (i.e., 12). The experimental results are shown in Figure~\ref{fig:speedup}, where it can be observed that, for all four datasets, DiCFS-hp scales better than DiCFS-vp. It can also be observed that the HIGGS and KDDCUP datasets are too small to take advantage of the use of more than two nodes and also that practically no speed-up improvement is obtained from increasing this value.

To summarize, our experiments show that even when vertical partitioning results in shorter execution times (the case in certain circumstances, e.g., when the dataset has an adequate number of features and instances for optimal parallelization according to the cluster resources), the benefits are not significant and may even be eclipsed by the effort invested in determining whether this approach is indeed the most efficient approach for a particular dataset or a particular hardware configuration or in  fine-tuning the number of partitions. Horizontal partitioning should therefore be considered as the best option in the general case. 

\begin{equation}
\label{eq:speedup}
speedup(m)=\left[ \frac { execution\quad time\quad on\quad 2\quad nodes }{execution\quad time\quad on\quad m\quad nodes }  \right] 
\end{equation}

We also compared the DiCFS-hp approach with that of Eiras-Franco et al.~\cite{Eiras-Franco2016}, who described a Spark-based distributed version of the CFS for regression problems. The comparison was based on their experiments with the HIGGS and EPSILON datasets but using our current hardware. Those datasets were selected as only having numerical features and so could naturally be treated as regression problems. Table~\ref{tbl:speedUp} shows execution time and speed-up values obtained for different sizes of both datasets for both distributed and non-distributed versions and considering them to be classification and regression problems. Regression-oriented versions for the Spark and WEKA versions are labelled RegCFS and RegWEKA, respectively, the number after the dataset name represents the sample size and the letter indicates whether the sample had removed or added instances (\emph{i}) or removed or added features (\emph{f}). In the case of oversized samples, the method used was the same as described above, i.e., features or instances were copied as necessary. The experiments were performed using ten cluster nodes for the distributed versions and a single node for the WEKA version. The resulting speed-up was calculated as the WEKA execution time divided by the corresponding Spark execution time.

The original experiments in~\cite{Eiras-Franco2016} were performed only using EPSILON\_50i and HIGGS\_100i. It can be observed that much better speed-up was obtained by the DiCFS-hp version for EPSILON\_50i but in the case of HIGGS\_100i, the resulting speed-up in the classification version was lower than the regression version. However, in order to have a better comparison, two more versions for each dataset were considered,  Table~\ref{tbl:speedUp} shows that the DiCFS-hp version has a better speed-up in all cases except in HIGGS\_100i dataset mentioned before.

\begin{table}[tbp]
\centering
\caption{Execution time and speed-up values for different CFS versions for regression and classification}
\label{tbl:speedUp}
\small
\begin{tabular}{lcccccc}
\hline
Dataset    & \multicolumn{4}{|c|}{Execution Time (sec)} & \multicolumn{2}{c}{Speed-Up} \\
           & \multicolumn{1}{|c}{WEKA} & \multicolumn{1}{c}{RegWEKA} & \multicolumn{1}{c}{DiCFS-hp} & \multicolumn{1}{c|}{RegCFS} & RegCFS       & DiCFS-hp      \\ \hline
EPSILON\_25i & 1011.42  & 655.56    & 58.85    & 63.61  & 10.31      & 17.19       \\
EPSILON\_25f & 393.91   & 703.95    & 25.83    & 55.08  & 12.78      & 15.25       \\
EPSILON\_50i & 4103.35  & 2228.64   & 76.98    & 110.13 & 20.24      & 53.30       \\
HIGGS\_100i  & 182.86   & 327.61    & 21.34    & 23.70  & 13.82      & 8.57        \\
HIGGS\_200i  & 2079.58  & 475.98    & 28.89    & 26.77  & 17.78      & 71.99       \\
HIGGS\_200f  & 934.07   & 720.32    & 21.42    & 34.35  & 20.97      & 43.61  \\
\hline      
\end{tabular}
\end{table}

\section{Conclusions and Future Work}
\label{sec:conclusions}

We describe two parallel and distributed versions of the CFS filter-based FS algorithm using the Apache Spark programming model: DiCFS-vp and DiCFS-hp. These two versions essentially differ in how the dataset is distributed across the nodes of the cluster. The first version distributes the data by splitting rows (instances) and the second version, following Ram\'irez-Gallego et al. ~\cite{Ramirez-Gallego2017}, distributes the data by splitting columns (features). As the outcome of a four-way comparison of DiCFS-vp and DiCFS-hp, a non-distributed implementation in WEKA and a distributed regression version in Spark, we can conclude as follows:

\begin{itemize}
\item As was expected, both DiCFS-vp and DiCFS-hp were able to handle larger datasets in much a more time-efficient manner than the classical WEKA implementation. Moreover, in many cases they were the only feasible way to process certain types of datasets because of prohibitive WEKA memory requirements.

\item Of the horizontal and vertical partitioning schemes, the horizontal version (DiCFS-hp) proved to be the better option in the general case due to its better scalability and its natural partitioning mode that enables the Spark framework to make better use of cluster resources.

\item For classification problems, the benefits obtained from distribution compared to non-distribution version can be considered equal to or even better than the benefits already demonstrated for the regression domain~\cite{Eiras-Franco2016}.
\end{itemize}

Regarding future research, an especially interesting line is whether it is necessary for this kind of algorithm to process all the data available or whether it would be possible to design automatic sampling procedures that could guarantee that, under certain circumstances, equivalent results could be obtained. In the case of the CFS, this question becomes more pertinent in view of the study of symmetrical uncertainty in datasets with up to 20,000 samples by Hall~\cite{Hall1999}, where tests showed that symmetrical uncertainty decreased exponentially with the number of instances and then stabilized at a certain number. Another line of future work could be research into different data partitioning schemes that could, for instance, improve the locality of data while overcoming the disadvantages of vertical partitioning.

\section*{Acknowledgements}
The authors thank CESGA for use of their supercomputing resources. This research has been partially supported by the Spanish Ministerio de Econom\'ia y Competitividad (research projects TIN 2015-65069-C2-1R, TIN2016-76956-C3-3-R), the Xunta de Galicia (Grants GRC2014/035 and ED431G/01) and the European Union Regional Development Funds. R. Palma-Mendoza holds a scholarship from the Spanish Fundaci\'on Carolina and the National Autonomous University of Honduras.

%\section*{References}

\bibliographystyle{elsarticle-num}

\begin{thebibliography}{10}
\expandafter\ifx\csname url\endcsname\relax
  \def\url#1{\texttt{#1}}\fi
\expandafter\ifx\csname urlprefix\endcsname\relax\def\urlprefix{URL }\fi
\expandafter\ifx\csname href\endcsname\relax
  \def\href#1#2{#2} \def\path#1{#1}\fi

\bibitem{Aha1991}
D.~W. Aha, D.~Kibler, M.~K. Albert, {Instance-Based Learning Algorithms},
  Machine Learning 6~(1) (1991) 37--66.
\newblock  \href
  {http://dx.doi.org/10.1023/A:1022689900470}
  {\path{doi:10.1023/A:1022689900470}}.

\bibitem{Bacardit2012}
J.~Bacardit, P.~Widera, A.~M{\'{a}}rquez-chamorro, F.~Divina, J.~S.
  Aguilar-Ruiz, N.~Krasnogor, {Contact map prediction using a large-scale
  ensemble of rule sets and the fusion of multiple predicted structural
  features}, Bioinformatics 28~(19) (2012) 2441--2448.
\newblock \href {http://dx.doi.org/10.1093/bioinformatics/bts472}
  {\path{doi:10.1093/bioinformatics/bts472}}.

\bibitem{bellman1957dynamic}
R.~Bellman, {{Dynamic
  Programming}}, Rand Corporation research study, Princeton University Press,
  1957.
%\newline\urlprefix\url{https://books.google.it/books?id=wdtoPwAACAAJ}

\bibitem{Bolon-Canedo2015a}
V.~Bol{\'{o}}n-Canedo, N.~S{\'{a}}nchez-Maro{\~{n}}o, A.~Alonso-Betanzos,
  {{Distributed
  feature selection: An application to microarray data classification}},
  Applied Soft Computing 30 (2015) 136--150.
\newblock \href {http://dx.doi.org/10.1016/j.asoc.2015.01.035}
  {\path{doi:10.1016/j.asoc.2015.01.035}}.
%\newline\urlprefix\url{http://linkinghub.elsevier.com/retrieve/pii/S156849461500054X}

\bibitem{Bolon-Canedo2015b}
V.~Bol{\'{o}}n-Canedo, N.~S{\'{a}}nchez-Maro{\~{n}}o, A.~Alonso-Betanzos,
  {Recent advances and emerging challenges of feature selection in the context
  of big data}, Knowledge-Based Systems 86 (2015) 33--45.
\newblock \href {http://dx.doi.org/10.1016/j.knosys.2015.05.014}
  {\path{doi:10.1016/j.knosys.2015.05.014}}.

\bibitem{Dash2003}
M.~Dash, H.~Liu,
  {{Consistency-based
  search in feature selection}}, Artificial Intelligence 151~(1-2) (2003)
  155--176.
\newblock \href {http://dx.doi.org/10.1016/S0004-3702(03)00079-1}
  {\path{doi:10.1016/S0004-3702(03)00079-1}}.
\newline\urlprefix\url{http://linkinghub.elsevier.com/retrieve/pii/S0004370203000791}

\bibitem{Dean2004a}
J.~Dean, S.~Ghemawat, {MapReduce: Simplied Data Processing on Large Clusters},
  Proceedings of 6th Symposium on Operating Systems Design and Implementation
  (2004) 137--149\href {http://arxiv.org/abs/10.1.1.163.5292}
  {\path{arXiv:10.1.1.163.5292}}, \href
  {http://dx.doi.org/10.1145/1327452.1327492}
  {\path{doi:10.1145/1327452.1327492}}.

\bibitem{Dean2008}
J.~Dean, S.~Ghemawat,
  {{MapReduce:
  Simplified Data Processing on Large Clusters}}, Communications of the ACM
  51~(1) (2008) 107.
\newline\urlprefix\url{http://dl.acm.org/citation.cfm?id=1327452.1327492}

\bibitem{duda2012pattern}
R.~O. Duda, P.~E. Hart, D.~G. Stork,
  {{Pattern Classification}}, John Wiley {\&} Sons, 2001.
%\newline\urlprefix\url{http://citeseerx.ist.psu.edu/viewdoc/download?doi=10.1.1.133.1318{\&}rep=rep1{\&}type=pdf}

\bibitem{Eiras-Franco2016}
C.~Eiras-Franco, V.~Bol{\'{o}}n-Canedo, S.~Ramos,
  J.~Gonz{\'{a}}lez-Dom{\'{i}}nguez, A.~Alonso-Betanzos, J.~Touri{\~{n}}o,
 {{Multithreaded
  and Spark parallelization of feature selection filters}}, Journal of
  Computational Science 17 (2016) 609--619.
\newblock \href {http://dx.doi.org/10.1016/j.jocs.2016.07.002}
  {\path{doi:10.1016/j.jocs.2016.07.002}}.
%\newline\urlprefix\url{http://linkinghub.elsevier.com/retrieve/pii/S1877750316301107}

\bibitem{Fayyad1993}
U.~M. Fayyad, K.~B. Irani,
  {{Multi-Interval
  Discretization of Continuos-Valued Attributes for Classification Learning}}
  (1993).
\newline\urlprefix\url{http://trs-new.jpl.nasa.gov/dspace/handle/2014/35171}

\bibitem{Garcia_aparallel}
D.~J. Garcia, L.~O. Hall, D.~B. Goldgof, K.~Kramer, {A Parallel Feature
Selection Algorithm from Random Subsets} (2004).

\bibitem{ghiselli1964theory}
E.~E. Ghiselli, {{Theory of
  Psychological Measurement}}, McGraw-Hill series in psychology, McGraw-Hill,
  1964.
\newline\urlprefix\url{https://books.google.es/books?id=mmh9AAAAMAAJ}

\bibitem{Guyon2003}
I.~Guyon, A.~Elisseeff, {An Introduction to Variable and Feature Selection},
  Journal of Machine Learning Research (JMLR) 3~(3) (2003) 1157--1182.
\newblock \href {http://arxiv.org/abs/1111.6189v1} {\path{arXiv:1111.6189v1}},
  \href {http://dx.doi.org/10.1016/j.aca.2011.07.027}
  {\path{doi:10.1016/j.aca.2011.07.027}}.

\bibitem{Hall1999}
M.~A. Hall, {Correlation-based feature selection for machine learning}, PhD
  Thesis., Department of Computer Science, Waikato University, New
  Zealand~(1999).
\newblock \href {http://dx.doi.org/10.1.1.37.4643} {\path{doi:10.1.1.37.4643}}.

\bibitem{Hall2000}
M.~A. Hall,
  {{Correlation-based
  Feature Selection for Discrete and Numeric Class Machine Learning}} (2000)
  359--366.
\newline\urlprefix\url{http://dl.acm.org/citation.cfm?id=645529.657793}

\bibitem{Hall2009a}
M.~Hall, E.~Frank, G.~Holmes, B.~Pfahringer, P.~Reutemann, I.~Witten, {The WEKA
  data mining software: An update}, SIGKDD Explorations 11~(1) (2009) 10--18.
\newblock \href {http://dx.doi.org/10.1145/1656274.1656278}
  {\path{doi:10.1145/1656274.1656278}}.

\bibitem{Ho1995}
T.~K. Ho, {{Random
  Decision Forests}}, in: Proceedings of the Third International Conference on
  Document Analysis and Recognition (Volume 1) - Volume 1, ICDAR '95, IEEE
  Computer Society, Washington, DC, USA, 1995, pp. 278----.
\newline\urlprefix\url{http://dl.acm.org/citation.cfm?id=844379.844681}

\bibitem{Idris2013}
A.~Idris, A.~Khan, Y.~S. Lee, {Intelligent churn prediction in telecom:
  Employing mRMR feature selection and RotBoost based ensemble classification},
  Applied Intelligence 39~(3) (2013) 659--672.
\newblock \href {http://dx.doi.org/10.1007/s10489-013-0440-x}
  {\path{doi:10.1007/s10489-013-0440-x}}.

\bibitem{Idris2012}
A.~Idris, M.~Rizwan, A.~Khan, {Churn prediction in telecom using Random Forest
  and PSO based data balancing in combination with various feature selection
  strategies}, Computers and Electrical Engineering 38~(6) (2012) 1808--1819.
\newblock \href {http://dx.doi.org/10.1016/j.compeleceng.2012.09.001}
  {\path{doi:10.1016/j.compeleceng.2012.09.001}}.

\bibitem{Kononenko1994}
I.~Kononenko,
  {{Estimating
  attributes: Analysis and extensions of RELIEF}}, Machine Learning: ECML-94
  784 (1994) 171--182.
\newblock \href {http://dx.doi.org/10.1007/3-540-57868-4}
  {\path{doi:10.1007/3-540-57868-4}}.
\newline\urlprefix\url{http://www.springerlink.com/index/10.1007/3-540-57868-4}

\bibitem{Kubica2011}
J.~Kubica, S.~Singh, D.~Sorokina,
  {{Parallel
  Large-Scale Feature Selection}}, in: Scaling Up Machine Learning, no.
  February, 2011, pp. 352--370.
\newblock \href {http://dx.doi.org/10.1017/CBO9781139042918.018}
  {\path{doi:10.1017/CBO9781139042918.018}}.
\newline\urlprefix\url{http://ebooks.cambridge.org/ref/id/CBO9781139042918A143}

\bibitem{Leskovec2014mining}
J.~Leskovec, A.~Rajaraman, J.~D. Ullman,
  {{Mining of Massive
  Datasets}}, 2014.
\newblock \href
  {http://dx.doi.org/10.1017/CBO9781139924801}
  {\path{doi:10.1017/CBO9781139924801}}.
\newline\urlprefix\url{http://ebooks.cambridge.org/ref/id/CBO9781139924801}

\bibitem{Lichman2013}
M.~Lichman, \href{http://archive.ics.uci.edu/ml}{{UCI Machine Learning
  Repository}} (2013).
\newline\urlprefix\url{http://archive.ics.uci.edu/ml}

\bibitem{Ma2009}
J.~Ma, L.~K. Saul, S.~Savage, G.~M. Voelker, {Identifying Suspicious URLs : An
  Application of Large-Scale Online Learning}, in: Proceedings of the
  International Conference on Machine Learning (ICML), Montreal, Quebec, 2009.

\bibitem{Palma-Mendoza2018}
R.~J. Palma-Mendoza, D.~Rodriguez, L.~De-Marcos,
  \href{http://link.springer.com/10.1007/s10115-017-1145-y}{{Distributed
  ReliefF-based feature selection in Spark}}, Knowledge and Information Systems
  (2018) 1--20\href {http://dx.doi.org/10.1007/s10115-017-1145-y}
  {\path{doi:10.1007/s10115-017-1145-y}}.
\newline\urlprefix\url{http://link.springer.com/10.1007/s10115-017-1145-y}

\bibitem{Peng2005}
H.~Peng, F.~Long, C.~Ding,
  {{Feature selection based
  on mutual information: criteria of max-dependency, max-relevance, and
  min-redundancy.}}, IEEE transactions on pattern analysis and machine
  intelligence 27~(8) (2005) 1226--38.
\newblock \href {http://dx.doi.org/10.1109/TPAMI.2005.159}
  {\path{doi:10.1109/TPAMI.2005.159}}.
\newline\urlprefix\url{http://www.ncbi.nlm.nih.gov/pubmed/16119262}

\bibitem{Peralta2015}
D.~Peralta, S.~del R{\'{i}}o, S.~Ram{\'{i}}rez-Gallego, I.~Riguero, J.~M.
  Benitez, F.~Herrera,
  {{Evolutionary
  Feature Selection for Big Data Classification: A MapReduce Approach
  }}, Mathematical Problems in Engineering 2015~(JANUARY).
\newblock \href {http://dx.doi.org/10.1155/2015/246139}
  {\path{doi:10.1155/2015/246139}}.
%\newline\urlprefix\url{http://sci2s.ugr.es/sites/default/files/2015-hindawi-peralta.pdf}

\bibitem{press1982numerical}
W.~H. Press, S.~A. Teukolsky, W.~T. Vetterling, B.~P. Flannery, {Numerical
  recipes in C}, Vol.~2, Cambridge Univ Press, 1982.

\bibitem{Quinlan1986}
J.~R. Quinlan, \href{http://dx.doi.org/10.1023/A:1022643204877}{{Induction of
  Decision Trees}}, Mach. Learn. 1~(1) (1986) 81--106.
\newblock \href {http://dx.doi.org/10.1023/A:1022643204877}
  {\path{doi:10.1023/A:1022643204877}}.
\newline\urlprefix\url{http://dx.doi.org/10.1023/A:1022643204877}

\bibitem{Quinlan1992}
J.~R. Quinlan,
  \href{http://portal.acm.org/citation.cfm?id=152181}{{C4.5:
  Programs for Machine Learning}}, Vol.~1, 1992.
\newblock \href {http://dx.doi.org/10.1016/S0019-9958(62)90649-6}
  {\path{doi:10.1016/S0019-9958(62)90649-6}}.
\newline\urlprefix\url{http://portal.acm.org/citation.cfm?id=152181}

\bibitem{Ramirez-Gallego2017}
S.~Ram{\'{i}}rez-Gallego, I.~Lastra, D.~Mart{\'{i}}nez-Rego,
  V.~Bol{\'{o}}n-Canedo, J.~M. Ben{\'{i}}tez, F.~Herrera, A.~Alonso-Betanzos,
  {{Fast-mRMR: Fast Minimum
  Redundancy Maximum Relevance Algorithm for High-Dimensional Big Data}},
  International Journal of Intelligent Systems 32~(2) (2017) 134--152.
\newblock \href {http://dx.doi.org/10.1002/int.21833}
  {\path{doi:10.1002/int.21833}}.
\newline\urlprefix\url{http://doi.wiley.com/10.1002/int.21833}

\bibitem{rish2001empirical}
I.~Rish, {An empirical study of the naive Bayes classifier}, in: IJCAI 2001
  workshop on empirical methods in artificial intelligence, Vol.~3, IBM, 2001,
  pp. 41--46.

\bibitem{Sadowski2014}
P.~Sadowski, P.~Baldi, D.~Whiteson, {Searching for Higgs Boson Decay Modes with
  Deep Learning}, Advances in Neural Information Processing Systems 27
  (Proceedings of NIPS) (2014) 1--9.

\bibitem{Silva2017}
J.~Silva, A.~Aguiar, F.~Silva,
  {{Parallel
  Asynchronous Strategies for the Execution of Feature Selection Algorithms}},
  International Journal of Parallel Programming (2017) 1--32\href
  {http://dx.doi.org/10.1007/s10766-017-0493-2}
  {\path{doi:10.1007/s10766-017-0493-2}}.
\newline\urlprefix\url{http://link.springer.com/10.1007/s10766-017-0493-2}

\bibitem{vapnik1995nature}
V.~Vapnik, {The Nature of Statistical Learning Theory} (1995).

\bibitem{Wang2016}
Y.~Wang, W.~Ke, X.~Tao, {{A Feature
  Selection Method for Large-Scale Network Traffic Classification Based on
  Spark}}, Information 7~(1) (2016) 6.
\newblock \href {http://dx.doi.org/10.3390/info7010006}
  {\path{doi:10.3390/info7010006}}.
\newline\urlprefix\url{http://www.mdpi.com/2078-2489/7/1/6}

\bibitem{XindongWu2014}
{Xindong Wu}, {Xingquan Zhu}, {Gong-Qing Wu}, {Wei Ding},
  \href{http://ieeexplore.ieee.org/document/6547630/}{{Data mining with big
  data}}, IEEE Transactions on Knowledge and Data Engineering 26~(1) (2014)
  97--107.
\newblock \href {http://dx.doi.org/10.1109/TKDE.2013.109}
  {\path{doi:10.1109/TKDE.2013.109}}.
\newline\urlprefix\url{http://ieeexplore.ieee.org/document/6547630/}

\bibitem{Zaharia2012}
M.~Zaharia, M.~Chowdhury, T.~Das, A.~Dave,
  {{Resilient
  distributed datasets: A fault-tolerant abstraction for in-memory cluster
  computing}}, NSDI'12 Proceedings of the 9th USENIX conference on Networked
  Systems Design and Implementation (2012) 2\href
  {http://arxiv.org/abs/EECS-2011-82} {\path{arXiv:EECS-2011-82}}, \href
  {http://dx.doi.org/10.1111/j.1095-8649.2005.00662.x}
  {\path{doi:10.1111/j.1095-8649.2005.00662.x}}.
%\newline\urlprefix\url{https://www.usenix.org/system/files/conference/nsdi12/nsdi12-final138.pdf}

\bibitem{Zaharia2010}
M.~Zaharia, M.~Chowdhury, M.~J. Franklin, S.~Shenker, I.~Stoica, {Spark :
  Cluster Computing with Working Sets}, HotCloud'10 Proceedings of the 2nd
  USENIX conference on Hot topics in cloud computing (2010) 10\href
  {http://dx.doi.org/10.1007/s00256-009-0861-0}
  {\path{doi:10.1007/s00256-009-0861-0}}.

\bibitem{Zhao2007}
Z.~Zhao, H.~Liu, {Searching for interacting features}, IJCAI International
  Joint Conference on Artificial Intelligence (2007) 1156--1161\href
  {http://dx.doi.org/10.3233/IDA-2009-0364} {\path{doi:10.3233/IDA-2009-0364}}.

\bibitem{Zhao2013a}
Z.~Zhao, R.~Zhang, J.~Cox, D.~Duling, W.~Sarle,
  {{Massively parallel
  feature selection: an approach based on variance preservation}}, Machine
  Learning 92~(1) (2013) 195--220.
\newblock \href {http://dx.doi.org/10.1007/s10994-013-5373-4}
  {\path{doi:10.1007/s10994-013-5373-4}}.
\newline\urlprefix\url{http://link.springer.com/10.1007/s10994-013-5373-4}

\end{thebibliography}

\end{document}